\documentclass{article}
\usepackage{preprint,times}
\usepackage{microtype}
\usepackage{amsmath,amssymb,amsthm,mathtools}
\usepackage{booktabs}
\usepackage{multirow}
\usepackage{algorithm}
\usepackage{algpseudocode}
\usepackage{placeins,flafter}
\usepackage{enumitem}
\usepackage{xcolor}
\usepackage{graphicx}
\usepackage{hyperref}
\usepackage{url}
\usepackage{tikz}
\usetikzlibrary{arrows.meta,positioning}

\hypersetup{
    colorlinks=true,
    linkcolor=blue,
    citecolor=blue,
    urlcolor=blue,
    pdftitle={TACIT-Switch: Cost-Aware Model Escalation for LLM Agents from Censored Supervision},
    pdfauthor={Ji'an Lei and Jian Huang}
}

\newcommand{\method}{\textsc{TACIT-Switch}}
\newcommand{\tacit}{\textsc{TACIT}}
\newcommand{\cheap}{\pi_{\mathrm{c}}}
\newcommand{\strong}{\pi_{\mathrm{s}}}

\newcommand{\event}{\kappa}
\newcommand{\E}{\mathbb{E}}
\newcommand{\Prob}{\mathbb{P}}
\newcommand{\ind}{\mathbb{I}}

\newtheorem{theorem}{Theorem}
\theoremstyle{definition}
\newtheorem{assumption}{Assumption}

\preprintfinalcopy

\title{TACIT-Switch: Cost-Aware Model Escalation\\
for LLM Agents from Censored Supervision}

\author{
Ji'an Lei\\
\normalfont School of Statistics\\
Beijing Normal University\\
Beijing 100875, China\\
\texttt{leijian2024@mail.bnu.edu.cn}
\And
Jian Huang\thanks{Corresponding author: \texttt{j.huang@polyu.edu.hk}}\\
\normalfont Department of Applied Mathematics\\
The Hong Kong Polytechnic University\\
Hung Hom, Kowloon, Hong Kong SAR, China\\
\texttt{j.huang@polyu.edu.hk}
}

\begin{document}
\maketitle

\begin{abstract}
When should a low-cost language-model agent hand an unfinished task to a
stronger model? Testing every possible handoff would require running the
stronger model from each intermediate state. We instead learn from paired
full-episode runs and an offline teacher that can mark a range of plausible
handoff steps without having to name one exact step. We treat these windows
as interval-censored observations of handoff timing. Our implementation, \method{}, fits a
mixture-cure threshold model to task features and trajectory diagnostics;
the deployed router requires no additional model call. In a multi-step
simulation it improves success by 7.4--11.1 percentage points over three
routing baselines at comparable cost. On 274 ALFWorld tasks, adaptive handoff
solved 159 tasks versus 137 for a development-selected fixed step at similar
mean weighted call cost; both policies let the stronger model continue from
the current state. On DABench, \method{} solved 49 of 67 tasks. Actual
continuations from teacher-marked states on ALFWorld frequently rescued
tasks that the cheaper model failed, but nearby checkpoints could work as well.
\end{abstract}

\section{Introduction}

Efficient agent deployment requires deciding when to use language-model
backbones with different capabilities and inference costs.  For a one-shot
request, routing is largely a static model-selection problem: choose a model
before generation begins.  In interactive agents, however, routing becomes a
sequential decision problem.  Each action alters the state from which later
decisions must proceed, while evidence of looping, protocol violations, or
stalled progress often emerges only as the trajectory unfolds.  Handing off
immediately to the larger model sacrifices the cost advantage of the smaller
model; handing off too late may leave little opportunity for recovery.
Routing therefore becomes a recoverability-aware stopping-time problem:
deciding whether and when to hand control to the larger model.

\begin{figure}[t]
\centering
\begin{tikzpicture}[
    cell/.style={draw,rounded corners=1pt,minimum width=0.64cm,
                 minimum height=0.43cm,font=\scriptsize},
    label/.style={anchor=east,font=\small\bfseries},
    note/.style={anchor=west,font=\scriptsize},
    >=Latex]
  \node[label] at (0,2.22) {Task routing};
  \node[cell,fill=gray!15] at (1.55,2.22) {route};
  \draw[->] (1.93,2.22) -- (2.25,2.22);
  \foreach \x in {2.6,3.4,4.2,5.0,5.8}
    \node[cell,fill=violet!12] at (\x,2.22) {\(M\)};
  \node[note] at (6.25,2.22) {one model throughout};

  \node[label] at (0,1.48) {SWE-Router};
  \node[cell,fill=blue!12] at (1.8,1.48) {C};
  \node[cell,fill=blue!12] at (2.6,1.48) {C};
  \node[cell,fill=blue!12] at (3.4,1.48) {C};
  \node[cell,fill=gray!15,minimum width=0.82cm] at (4.18,1.48) {value};
  \draw[->] (4.64,1.48) -- (4.88,1.74);
  \draw[->] (4.64,1.48) -- (4.88,1.22);
  \node[cell,fill=blue!12,minimum width=1.12cm,minimum height=0.36cm]
    at (5.48,1.74) {C continue};
  \node[cell,fill=red!15,minimum width=1.12cm,minimum height=0.36cm]
    at (5.48,1.22) {S restart};
  \node[note] at (6.25,1.48) {one decision after \(K\)};

  \node[label] at (0,0.74) {ReDAct};
  \node[cell,fill=blue!12] at (1.8,0.74) {C};
  \node[cell,fill=blue!12] at (2.6,0.74) {C};
  \node[cell,fill=red!15]  at (3.4,0.74) {S};
  \node[cell,fill=blue!12] at (4.2,0.74) {C};
  \node[cell,fill=red!15]  at (5.0,0.74) {S};
  \node[cell,fill=blue!12] at (5.8,0.74) {C};
  \node[note] at (6.25,0.74) {step-local; C resumes};

  \node[label] at (0,0) {\method};
  \node[cell,fill=blue!12] at (1.8,0) {C};
  \node[cell,fill=blue!12] at (2.6,0) {C};
  \node[cell,fill=blue!12] at (3.4,0) {C};
  \draw[dashed,thick] (3.8,-0.32) -- (3.8,0.32);
  \node[cell,fill=red!15] at (4.2,0) {S};
  \node[cell,fill=red!15] at (5.0,0) {S};
  \node[cell,fill=red!15] at (5.8,0) {S};
  \node[note] at (6.25,0) {adaptive, one-way};
  \draw[->,blue!65!black] (1.55,-0.48) -- (3.72,-0.48)
    node[midway,below,font=\scriptsize] {accumulated risk};
\end{tikzpicture}
\caption{Four model-routing strategies distinguished by when they make routing
decisions: task routing (one pre-execution model choice), SWE-Router (one
post-prefix decision to continue with \(\mathrm{C}\) or restart with
\(\mathrm{S}\)), ReDAct (step-local,
non-permanent deferral), and \mbox{\method{}} (an adaptive one-way handoff from the
current trajectory).  Here \(\mathrm{C}\) and \(\mathrm{S}\) denote Cheap and
Strong, respectively.}
\label{fig:motivation}
\end{figure}
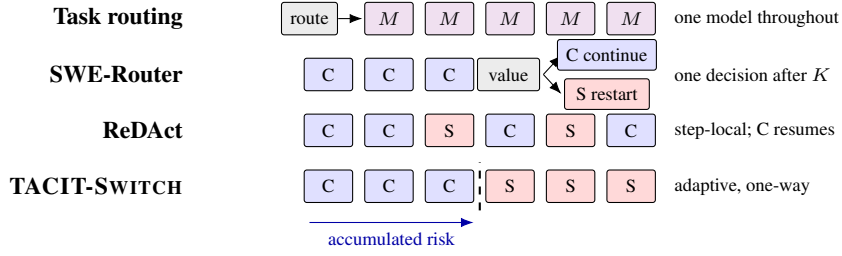

Existing methods make routing decisions at different points in execution
(Figure~\ref{fig:motivation}).  Request-level methods do so without observing
an evolving environment trajectory: RouteLLM chooses a model before generation,
whereas FrugalGPT uses a learned cascade
~\citep{ong2025routellm,chen2024frugalgpt}.  ReDAct and TRIM make step-local,
non-permanent interventions after which control returns to the smaller model
~\citep{redact2026,kapoor2026trim}; SWE-Router makes one
continue-or-restart decision after a fixed prefix~\citep{son2026swerouter}.
EvoRoute instead repeatedly selects models across explicit workflow
subtasks~\citep{zhang2026evoroute}.  We study a different control structure:
an adaptively timed, one-way transfer for the remainder of the episode, which
we call a \emph{permanent handoff}.

\FloatBarrier

TRIM assumes delimited reasoning steps and step-level correctness scores,
whereas EvoRoute assumes a role-structured multi-model workflow and an
experience base.  Neither method maps directly to our fixed two-model setting,
so we compare task-, step-, and fixed-prefix routing policies under the same
protocol.

The difficulty is obtaining training labels. A Cheap trajectory tells us what
happened while Cheap stayed in control, not what Strong would have done from
the same state. Estimating that difference at every step would require many
additional Strong continuations. An offline teacher can instead inspect
existing runs. Unlike retrospective failure attribution, which asks which
step caused an error~\citep{zhang2025attribution}, our question is when a
handoff might still help. Suppose the teacher judges steps 6--8 plausible but
cannot distinguish step 7 from its neighbors. Training on step 7 alone would
treat that uncertainty as a precise observation.

We call our supervision \emph{Teacher-Annotated Censored Intervention Times}
(\tacit{}). We run Cheap and Strong once each from the start; when Cheap fails
and Strong succeeds, the teacher marks a plausible handoff window on the
Cheap trajectory. We model the window as an interval-censored observation,
not as a verified optimal intervention time. This use of a teacher differs
from observing when an expert actually intervened~\citep{spencer2020interventions}.
What is new here is using a teacher's uncertainty about \emph{when} to
intervene as interval-censored supervision for permanent handoff between
LLM agents. The annotation protocol does not prescribe a particular timing
model: a parametric or nonparametric learner could be used if it can fit
the censored observations.

We separate two uncertainties: whether the paired Strong rollout succeeds and,
conditional on that outcome, where the handoff threshold lies on the
cumulative-risk scale.  This incidence--threshold decomposition adapts the
logic of mixture-cure models~\citep{farewell1982mixture,kuk1992mixture}.
Coarse teacher intervals are represented as interval-censored observations
~\citep{turnbull1976interval}, and a log-normal AFT-style parameterization
models the positive, task-dependent threshold
\(\kappa_i\)~\citep{wei1992aft}.  The corresponding handoff step is not modeled
directly; it is induced by the first crossing,
\(\tau_i^\star=\inf\{t:R_i(t)\geq\kappa_i\}\).
Under this working factorization,
\begin{equation}
q_i(t)=\pi_\gamma(x_i)F_{\beta,s}\!\left(R_i(t)\mid x_i\right)
\label{eq:intro-score}
\end{equation}
is the product of the estimated probability that the paired Strong rollout
succeeds and the conditional probability that its threshold has been crossed.
The policy hands control to Strong when this score reaches a threshold
selected on development data. Online, it uses only task descriptors and
diagnostics from Cheap's own outputs: no teacher, value head, or separate
routing LLM is called. Strong's chance of success from the initial state
bounds the \emph{score} by construction, not the success of a policy that
hands Strong a partially completed task.

We study the estimator's local sensitivity to annotation error
(Theorem~\ref{thm:supervision-robustness}), then evaluate routing in
simulation and on two interactive benchmarks. On ALFWorld we separately
test whether adaptive timing improves on fixed-step continuation and whether
Strong can actually finish from teacher-marked states.

\section{Problem Formulation and TACIT Supervision}
\label{sec:problem}

\subsection{Permanent handoff on a cumulative-risk scale}

We consider episodes of at most \(H\) interaction steps.  A permanent-handoff
policy \(h\) begins each episode with a Cheap policy \(\cheap\).  Based on the
trajectory observed so far, \(h\) may transfer control once to a Strong policy
\(\strong\); after transfer, \(\strong\) produces all remaining actions.  Let
\(Y_i(h)\in\{0,1\}\) and
\(\mathrm{Cost}_i(h)\) denote task success and deployment inference cost.

We seek policies that maximize success under an inference budget,
\begin{equation}
  \max_h\;\E[Y_i(h)] \qquad \text{s.t.}\qquad \E[\mathrm{Cost}_i(h)]\le C_{\max}.
  \label{eq:budget-objective}
\end{equation}
This constraint defines the success--cost trade-off.  Here \(Y_i(h)\) records
task-specific performance, while \(\mathrm{Cost}_i(h)\) is defined by the
deployment setting and may vary across tasks.  \method{} does not require a
particular cost model.  It first estimates the handoff score
\(\widehat q_i(t)\); the application-specific cost enters only when choosing
its decision threshold \(\alpha\) on development data.  This threshold is then
fixed for evaluation and deployment.

To determine when to hand off from the
observed trajectory, we represent the handoff threshold on a learned
cumulative-risk scale rather than at a fixed time step.  Specifically, at
each decision checkpoint \(t\), we map the observed trajectory prefix
\(\mathcal H_{i,t}\) to a nonnegative diagnostic vector
\(u_i(t)=\phi(\mathcal H_{i,t})\in\mathbb R_+^d\).  Its entries capture signals
such as repeated or invalid actions, uncertainty, and stalled progress.  We
learn nonnegative diagnostic weights \(w=(w_1,\ldots,w_d)\) satisfying
\(\sum_{j=1}^d w_j=1\).  The stepwise and cumulative risk scores are
\begin{equation}
  r_i(t)=w^\top u_i(t),\qquad R_i(t)=\sum_{s\le t}r_i(s).
  \label{eq:cum-risk}
\end{equation}
Because both \(u_i(t)\) and \(w\) are nonnegative, \(R_i(t)\) is nondecreasing
as evidence accumulates.
For an episode whose paired Strong rollout succeeds, let
\(\event_i>0\) denote its latent handoff threshold on the cumulative-risk
scale.  The corresponding first-crossing step is
\begin{equation}
  \tau_i^\star=\inf\{t:R_i(t)\ge\event_i\},
  \qquad \inf\varnothing=\infty.
  \label{eq:latent-crossing}
\end{equation}
This crossing is a decision-model construct; it does not change either model's
ability or the environment dynamics.

\subsection{Censored TACIT supervision}

Each training task provides paired final outcomes from complete Cheap and
Strong rollouts.  When Cheap fails but Strong succeeds, an offline teacher
additionally marks a coarse interval in which handoff appears useful and still
recoverable.  Let \(A_i\) and \(B_i\) indicate Cheap and Strong success,
respectively, from the initial state, and abbreviate the outcome pair
\((A_i,B_i)\) as \(AB\).  Thus \(B_i\) records the realized outcome of one
paired Strong rollout from the initial state; it is not an intrinsic or
state-independent capability label.
In the working likelihood, records with \(B_i=0\) enter the cure component.
This is a modeling convention: \(B_i=0\) does not imply that Strong could
never finish from a state reached by Cheap. Records with \(B_i=1\) form
the paired-Strong-success stratum and use a finite handoff-threshold
distribution.
Table~\ref{tab:ab-types} summarizes the resulting evidence.

\begin{table}[H]
\caption{Mapping from paired full-episode rollout outcomes to handoff
evidence.  Only \(AB=01\) receives a teacher interval.}
\label{tab:ab-types}
\centering
\small
\begin{tabular}{cll}
\toprule
Type & Paired outcomes & Likelihood evidence \\
\midrule
01 & cheap fails, strong succeeds & paired-Strong success; interval (or time missing) \\
11 & both succeed & paired-Strong success; right censoring \\
10 & cheap succeeds, strong fails & operational cure component \\
00 & both fail & operational cure component \\
\bottomrule
\end{tabular}
\end{table}

For \(AB=01\), an offline teacher reads the complete Cheap trajectory and marks
either a post-action checkpoint \(\widetilde{\tau}_i\) or a wider range of
checkpoints at which handoff first appears useful and recovery remains
feasible.  A single marked checkpoint does not identify an exact real-valued
threshold; it gives the one-step interval
\begin{equation}
  R_i(\widetilde{\tau}_i-1)<\event_i\le R_i(\widetilde{\tau}_i).
  \label{eq:tacit-interval}
\end{equation}
More generally, a teacher range \([a_i,b_i]\) gives
\(R_i(a_i-1)<\event_i\le R_i(b_i)\), with the convention \(R_i(-1)=0\).
These interval-censored observations constitute \tacit{} supervision.  The
teacher coarsens only the threshold location; membership in the
paired-Strong-success stratum already follows from \(B_i=1\).  \(AB=11\)
belongs to this stratum but is right-censored because Cheap completes the
episode without an observed need for rescue.

\section{TACIT-Switch: Model, Estimation, and Deployment}
\label{sec:model}

\subsection{Incidence--threshold model}

We choose a parametric model whose terms can be interpreted separately.
The incidence--threshold model combines three established survival-analysis
tools: mixture-cure models separate susceptibility from the conditional event
distribution~\citep{farewell1982mixture,kuk1992mixture}, AFT models regress
log event times on covariates~\citep{wei1992aft}, and interval censoring
represents event locations known only through a window
~\citep{turnbull1976interval}.  In our application, \(B_i\) is the observed
incidence outcome, and the event location is a handoff threshold on learned
cumulative risk rather than elapsed time. A discrete-time hazard model can
learn from the same windows without positing a threshold on cumulative risk
(Appendix~\ref{app:control-study}).

Accordingly, a logistic component models the observed Strong-success
indicator \(B_i\), while a log-normal AFT component models the threshold
location conditional on \(B_i=1\):
\begin{equation}
  \pi_i=\Prob(B_i=1\mid x_i)=\operatorname{sigmoid}(x_i^\top\gamma),\qquad
  \log\event_i\mid (B_i=1)=x_i^\top\beta+s\epsilon_i,
  \quad\epsilon_i\sim\mathcal{N}(0,1).
  \label{eq:mixture-aft}
\end{equation}
Here \(x_i\) contains compact task features and \(F_{\beta,s}(\cdot\mid x_i)\)
denotes the conditional CDF of a finite threshold.

\subsection{Censored likelihood and estimation}

Write \(R_{i,w}\) for the cumulative score in
Equation~\eqref{eq:cum-risk} to emphasize its dependence on \(w\).  For a
teacher window \([a_i,b_i]\), define
\(L_i(w)=R_{i,w}(a_i-1)\) and \(U_i(w)=R_{i,w}(b_i)\), with
\(R_{i,w}(-1)=0\).  For a right-censored episode ending at \(H_i\), define
\(C_i(w)=R_{i,w}(H_i)\).  Because \(B_i\) is observed for every paired rollout,
the main working-likelihood contributions are
\begin{equation}
\mathcal{L}_i(\theta)=
\begin{cases}
1-\pi_i, & \text{cured }(B_i=0),\\
\pi_i\!\left[F_{\beta,s}(U_i(w)\mid x_i)-F_{\beta,s}(L_i(w)\mid x_i)\right],
  & \text{teacher interval }(B_i=1),\\
\pi_i\!\left[1-F_{\beta,s}(C_i(w)\mid x_i)\right],
  & \text{right-censored }(B_i=1).
\end{cases}
\label{eq:working-likelihood}
\end{equation}
If \(B_i=1\) but no teacher time is available, marginalizing over the
unobserved threshold location leaves the incidence contribution \(\pi_i\).
We jointly estimate \(\theta=(w,\gamma,\beta,s)\) by maximizing the working
likelihood with \(\ell_2\) regularization on the non-intercept coefficients of
\(\gamma\) and \(\beta\).  Appendix~\ref{app:likelihood} gives the objective
and inference pseudocode.
When the working model and its coarsening assumptions hold,
Equation~\eqref{eq:working-likelihood} is the observed-data likelihood.  For
agent data, however, the teacher constructs the interval after inspecting the
realized trajectory, so the coarsening
mechanism need not be ignorable~\citep{heitjan1991coarse}.  We therefore use
Equation~\eqref{eq:working-likelihood} as a working likelihood for learning the
switching policy, without claiming a generative model for the trajectory or
teacher annotation.

For finite-sample analysis, we use unconstrained local coordinates for the
simplex weight \(w\) and positive scale \(s\), and let \(p\) be the number of
free parameters.  Let \(\theta^\dagger\) denote a local minimizer of the clean
population version of the regularized working objective.  Assumption~1 defines
the neighborhood radius \(r\), local curvature \(\mu\), score-concentration
scale \(\sigma\), and interval-corruption influence bound \(G_T\).

\begin{theorem}[Local stability under teacher-interval noise]
\label{thm:supervision-robustness}
Fix \(\delta\in(0,1)\).  Under Assumption~\ref{ass:local-regularity},
suppose that teacher-provided intervals are incorrect for at most a fraction
\(\rho_T\) of the \(n\) records.  Define
\begin{equation}
  a_n(\delta)
  =2\sigma\sqrt{\frac{2\{p\log 5+\log(4/\delta)\}}{n}}
   +G_T\rho_T.
  \label{eq:local-radius}
\end{equation}
If \(a_n(\delta)<\mu r/2\), then, with probability at least \(1-\delta\),
the corrupted empirical objective has a unique stationary point
\(\widehat\theta\in\mathbb B_r(\theta^\dagger)\), and
\begin{equation}
  \|\widehat\theta-\theta^\dagger\|_2
  \le \frac{2a_n(\delta)}{\mu}.
  \label{eq:supervision-robustness}
\end{equation}
For a task with features \(x\) and observed diagnostics \(u(1{:}t)\), let
\(q_\theta(x,u(1{:}t))=\pi_\gamma(x)F_{\beta,s}(R_w(t)\mid x)\) denote the
model's handoff score at step \(t\).  If this score is uniformly
\(L_q\)-Lipschitz in \(\theta\) over all evaluation tasks, decision steps,
and \(\mathbb B_r(\theta^\dagger)\), then on the same event, for every such
task and step,
\begin{equation}
  |q_{\widehat\theta}(x,u(1{:}t))
    -q_{\theta^\dagger}(x,u(1{:}t))|
  \le \frac{2L_q a_n(\delta)}{\mu}.
  \label{eq:score-recovery}
\end{equation}
\end{theorem}
Thus, relative to the clean population working target, parameter and
handoff-score errors are \(O\!\left(\sqrt{p/n}+\rho_T\right)\), where the two
terms capture finite-sample variation and teacher-interval corruption.
Appendix~\ref{app:finite-sample-proof} states the assumptions and gives the
full proof.

\subsection{Cost-sensitive online handoff}

At deployment, the teacher is absent.  At each benchmark-defined decision
checkpoint, the policy updates \(R_i(t)\) from the Cheap trajectory evidence
available at that checkpoint and computes the fitted handoff score
\begin{equation}
  \widehat q_i(t)=\hat\pi_iF_{\hat\beta,\hat s}(R_i(t)\mid x_i).
  \label{eq:handoff-score}
\end{equation}
It permanently hands control to \(\strong\) when
\begin{equation}
  \widehat q_i(t)\ge\alpha.
  \label{eq:decision-rule}
\end{equation}

Because the implemented diagnostics are nonnegative, the cumulative score is
nondecreasing.  The deployed switch time is therefore the first decision
checkpoint satisfying Equation~\eqref{eq:decision-rule}, with no switch if the
set is empty.  Lowering \(\alpha\) cannot delay this crossing.

We select \(\alpha\) on development data by maximizing empirical success under
the prescribed cost budget and freeze it before testing. A systematic
difference between teacher and fitted handoff times can also be checked on
development data: shift the training windows by a common offset, refit, and
select the offset and threshold under the same budget. Zero offset is a
candidate, not a correction that must be made (Appendix~\ref{app:window-study}).

\section{Controlled Statistical Experiments}
\label{sec:controlled-experiments}

\begin{figure*}[t]
\centering
\includegraphics[width=\textwidth]{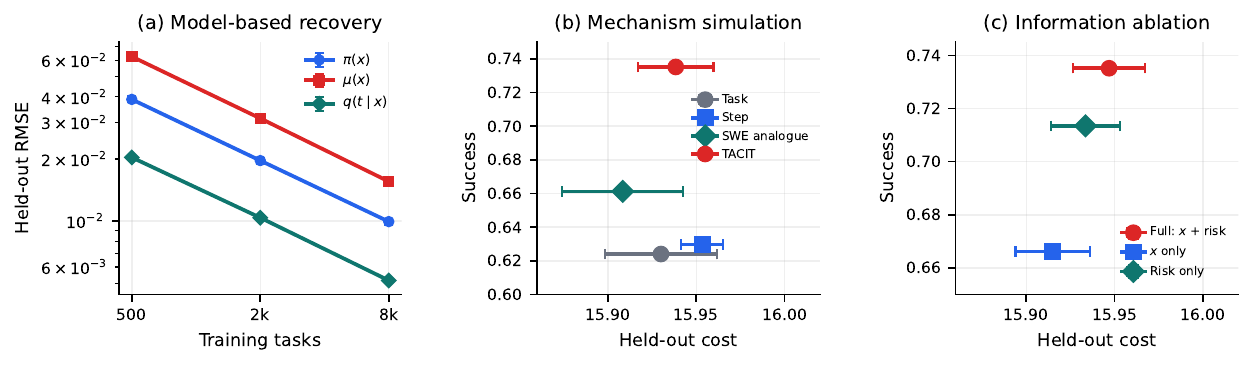}
\caption{Controlled recovery, policy, and ablation results.
(a) Held-out recovery error over 500 independent fits per sample size.
(b) The 100-repeat mechanism-based policy comparison, including the
development-tuned SWE-Router analogue.
(c) The 100-repeat information ablation.
Points show replicate means; bars show Student-\(t\) 95\% Monte Carlo
confidence intervals for those means.
Candidate policies are fitted on training data, selected on development data, and
evaluated once on test after all choices are frozen.}
\label{fig:synthetic-core}
\end{figure*}

We use two complementary simulations.  The first generates censored
observations directly from the proposed AB--AFT model and tests whether its
parameters and predictions can be recovered.  The second generates complete
multi-step trajectories from an agent-like path-state process and tests
whether the fitted switching policy improves success at a fixed cost.  This
second simulation is not assumed to follow the AB--AFT model and therefore
evaluates decision utility rather than parameter recovery.  Information
ablations and one-factor sensitivity experiments then isolate the roles of
task features and online risk.
Figures~\ref{fig:synthetic-core} and~\ref{fig:synthetic-sensitivity} summarize
the four statistical studies.
Throughout this section, ``\(\pm\)'' denotes the half-width of a Student-\(t\)
95\% Monte Carlo confidence interval across independent replicates.

\subsection{Model recovery under correct specification}

We first isolate the statistical behavior of the estimator under correct model
specification.  Data are generated according to
\[
B_i\mid x_i\sim\mathrm{Bernoulli}(\pi_i),\qquad
\log\kappa_i=\widetilde x_i^\top\beta^\star+s^\star\varepsilon_i,
\quad \varepsilon_i\sim N(0,1).
\]
Writing \(\mu(x)=\widetilde x^\top\beta^\star\), the induced first-crossing
time and handoff probability are
\[
T_i=\inf\{t:R_i(t)\ge\kappa_i\},\qquad
q(t\mid x_i)=\pi(x_i)\Pr\{\kappa_i\le R_i(t)\mid x_i,B_i=1\}.
\]
For \(B_i=1\), a crossing by the horizon produces an AB01 interval, whereas
\(T_i>H\) yields an AB11 right-censored observation.  When
\(B_i=0\), a separate Bernoulli nuisance model for \(A_i\) yields AB00 or
AB10.  We fit 500 independent samples at each
\(n\in\{500,2{,}000,8{,}000\}\) and evaluate every fit on a fresh sample of
10,000 tasks.

All 1,500 fits converge.  Componentwise parameter RMSE approximately halves
whenever the sample size quadruples, while held-out \(q(t\mid x)\) RMSE
decreases from \(0.0203\pm0.0006\) at \(n=500\) to
\(0.0052\pm0.0001\) at \(n=8{,}000\).  These results are consistent with an
\(n^{-1/2}\)-like recovery rate.  The mean AB00/01/10/11 proportions remain
stable at \(0.259/0.261/0.197/0.284\).

Appendix~\ref{app:noise-robustness} studies two types of corruption at rates up
to \(20\%\): moving a one-step teacher interval to an adjacent step and
flipping the paired Strong outcome.  The estimator changes little under the
former but is more sensitive to the latter.

\subsection{Mechanism-based multi-step simulation}

To test policy utility outside the fitted likelihood, we use a separate
ten-step, three-state simulator in which a task succeeds after at least eight
correct actions.  Cheap and Strong follow fixed state-transition and
action-success probabilities, with Strong more likely to recover from
degraded states.  A noisy post-step signal provides partial evidence about
path quality, while \(x\) provides a weak task-level prior.  The latent task
threshold affects only synthetic supervision and diagnostic quantities; it
does not enter actor dynamics, risk emissions, or terminal success.

Across 100 independent replicates, policies are fitted on 8,000 training tasks,
tuned on 4,000 development tasks under a cost cap of 16, and evaluated on
20,000 test tasks after all choices are frozen.  Cheap and Strong steps cost 1
and 3.  We compare Task Router, Step Deferral using the latest risk, a
fixed-prefix SWE-Router analogue~\citep{son2026swerouter}, and \method{};
implementation and teacher details are deferred to
Appendix~\ref{app:protocol}.

\noindent\begin{minipage}{\linewidth}
Pure Cheap and Pure Strong succeed on \(44.59\%\) and \(84.88\%\) of test
tasks.  At nearly identical realized costs (\(15.91\)--\(15.95\)), Task
Router, Step Deferral, the SWE-Router analogue, and \method{} attain
\(62.40\%\), \(62.98\%\), \(66.13\%\), and \(73.52\%\) success.  The paired
gains of \method{} are \(11.12\), \(10.54\), and \(7.39\) percentage points,
all positive in every replicate.  The gain comes from using trajectory
evidence to identify when Strong is more useful.
\end{minipage}

\subsection{Information ablation}

The path-state construction makes task information and trajectory information
individually imperfect, allowing us to measure their separate contributions.
On the same 100-replicate splits, we compare the Full model with \(x\)-only and
risk-only variants.  The \(x\)-only variant masks the diagnostic values but
retains indicators of which Cheap steps were executed.

At nearly identical realized cost, Full reaches \(73.51\pm0.09\%\) success,
compared with \(66.62\pm0.10\%\) for \(x\)-only and
\(71.35\pm0.09\%\) for risk-only.  Its paired gains are
\(6.90\pm0.08\) and \(2.17\pm0.08\) percentage points, respectively.
Online risk provides the larger marginal gain, while the improvement over
risk-only shows that task features add information beyond online risk.

\subsection{Scenario sensitivity}

Finally, we vary task-feature and online-risk information one at a time,
holding actor behavior and endpoint difficulty fixed.  Each axis contains five
settings spanning the observed AUC range under the same development-only
selection protocol; full results appear in
Appendix~\ref{app:protocol}.

\begin{figure}[H]
\centering
\includegraphics[width=0.88\textwidth]{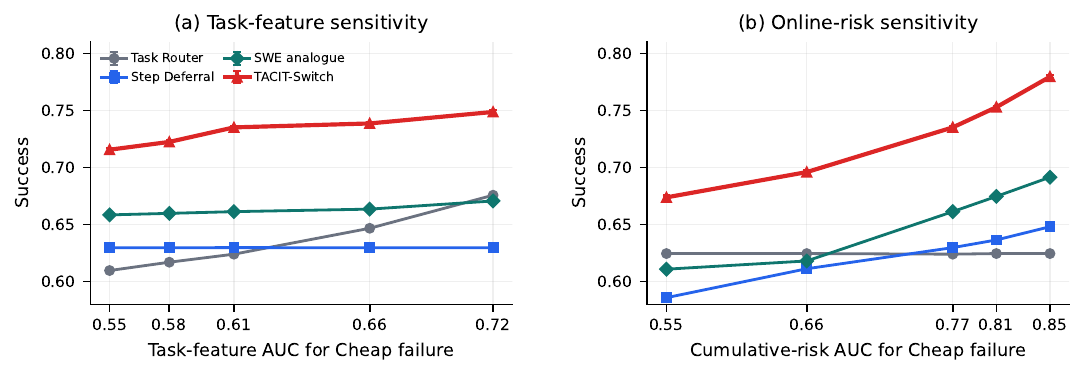}
\caption{One-factor information sensitivity.
(a) Task-feature information is varied while the execution process is fixed;
the horizontal axis is the AUC with which task features predict terminal Cheap
failure.  (b) Online-risk information is varied by changing only risk-emission
noise; the horizontal axis is the corresponding AUC for cumulative risk.
Both panels compare Task Router, Step Deferral, the SWE-Router analogue, and
\mbox{\method{}}.  The baseline uses 100 repeats and each nonbaseline
condition uses 50.  Points show replicate means; bars show Student-\(t\) 95\%
Monte Carlo confidence intervals for those means.}
\label{fig:synthetic-sensitivity}
\end{figure}

Figure~\ref{fig:synthetic-sensitivity} shows that \method{} has higher success
than all baselines at comparable cost.  Success increases monotonically across
the five task-feature settings, from \(71.56\%\) to \(74.87\%\), and across the
five online-risk settings, from \(67.38\%\) to \(77.96\%\).
Pure-Cheap and Pure-Strong success remain stable, so these trends reflect
information quality rather than changes in task difficulty or actor
capability.

\section{Interactive-Agent Experiments}
\label{sec:interactive-agent-experiments}

\begin{figure}[!b]
\centering
\includegraphics[width=\linewidth]{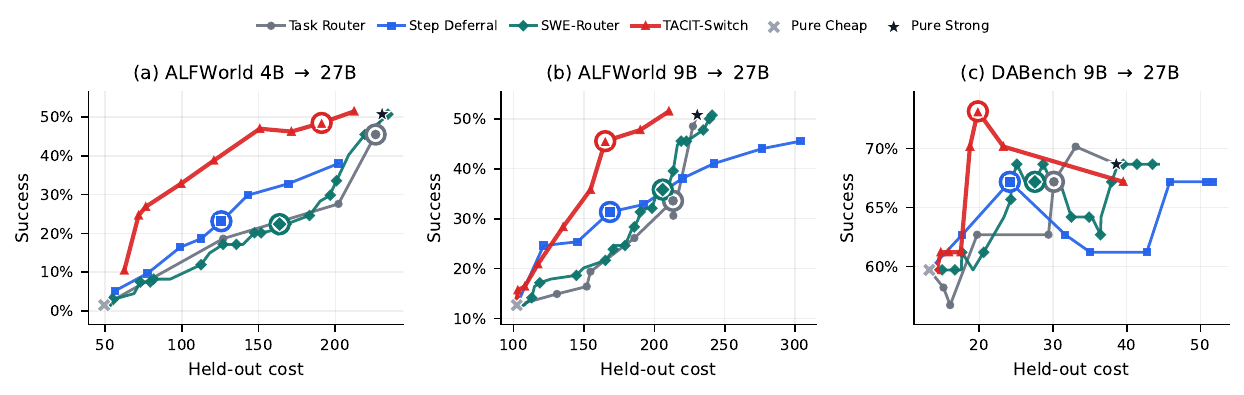}
\caption{\textbf{Held-out success--cost curves for the real-agent
experiments.}  Lines trace test performance over the prespecified policy
grids; rings mark the operating points selected on development data.  The
remaining test points show curve shape only and were not used for selection.}
\label{fig:real-agent-pareto}
\end{figure}

\subsection{Common protocol}

Within each benchmark, Cheap and Strong share the scaffold, prompt, action
space, and step budget. For the primary policies, DeepSeek V4 Pro annotates
training trajectories on which Cheap failed and Strong succeeded from the
start. The smaller studies use the recommended checkpoint as a one-step
interval; the larger ALFWorld study retains the full range. No policy queries
the teacher during execution. Separate annotations used to compare teacher
and policy timing are for analysis, not inputs to the deployed router.

We compare Pure Cheap and Pure Strong, Task Router, a ReDAct-inspired
step-local deferral baseline~\citep{redact2026}, a SWE-Router-style fixed-prefix
baseline~\citep{son2026swerouter}, and \method{}.  All policy settings are
chosen once on development data under a fixed cost cap and frozen before test.
The cost proxy counts calls weighted by model size in 4B-equivalent units,
not tokens, latency, or fees. It is used for operating-point selection, not
for fitting the score.
Table~\ref{tab:main-results} and
Figure~\ref{fig:real-agent-pareto} report the test results;
Appendix~\ref{app:runtime-study} also accounts for the tokens Strong reads
when it inherits Cheap's history.

\begin{table}[!b]
\caption{\textbf{Frozen interactive-agent operating points.}
All models are fitted on training data.  Operating points and policy
hyperparameters, including the SWE-Router prefix length and threshold, are
selected on development data and frozen before test.
Strong share is the fraction of deployed model calls made by Strong.
\(\dagger\) marks the proposed method.}
\label{tab:main-results}
\centering
\small
\begin{tabular}{llrrr}
\toprule
Benchmark & Policy & Success & Avg. cost & Strong share \\
\midrule
\multirow{6}{*}{DABench Test67}
 & Pure 9B & 40/67 (59.7\%) & 13.23 & 0.0\% \\
 & Task router & 45/67 (67.2\%) & 30.12 & 68.3\% \\
 & Step deferral & 45/67 (67.2\%) & 24.15 & 22.9\% \\
 & SWE-Router (\(K=2\)) & 45/67 (67.2\%) & 27.55 & 47.3\% \\
 & \method{}\(^{\dagger}\) & 49/67 (73.1\%) & 19.81 & 33.3\% \\
 & Pure 27B & 46/67 (68.7\%) & 38.59 & 100.0\% \\
\midrule
\multirow{6}{*}{ALFWorld unseen134 (4B)}
 & Pure 4B & 2/134 (1.5\%) & 49.40 & 0.0\% \\
 & Task router & 61/134 (45.5\%) & 226.26 & 92.3\% \\
 & Step deferral & 31/134 (23.1\%) & 125.79 & 22.4\% \\
 & SWE-Router (\(K=2\)) & 30/134 (22.4\%) & 163.72 & 46.9\% \\
 & \method{}\(^{\dagger}\) & 65/134 (48.5\%) & 191.10 & 79.5\% \\
 & Pure 27B & 68/134 (50.7\%) & 230.51 & 100.0\% \\
\midrule
\multirow{6}{*}{ALFWorld unseen134 (9B)}
 & Pure 9B & 17/134 (12.7\%) & 102.22 & 0.0\% \\
 & Task router & 45/134 (33.6\%) & 213.26 & 72.8\% \\
 & Step deferral & 42/134 (31.3\%) & 168.46 & 22.8\% \\
 & SWE-Router (\(K=4\)) & 48/134 (35.8\%) & 205.95 & 60.3\% \\
 & \method{}\(^{\dagger}\) & 61/134 (45.5\%) & 165.11 & 47.3\% \\
 & Pure 27B & 68/134 (50.7\%) & 230.51 & 100.0\% \\
\bottomrule
\end{tabular}
\end{table}

\subsection{ALFWorld}

ALFWorld is a text-based interactive household-task benchmark aligned with
embodied environments~\citep{shridhar2021alfworld}.  We pair Qwen3.5-4B or
Qwen3.5-9B with Qwen3.6-27B, allow at most 50 actions, fit on 100 valid-seen
episodes, select on 40 held-out seen episodes, and evaluate once on 134
valid-unseen episodes.  All operating settings are chosen on the development
split; Appendix~\ref{app:protocol} gives the outcome counts and candidate
grids.

Table~\ref{tab:main-results} and Figure~\ref{fig:real-agent-pareto} show that
\method{} has the highest success among the learned policies in the table for both
model pairs.  With 9B Cheap, it achieves higher success at lower cost than both
Task Router and SWE-Router; with 4B Cheap, it approaches pure-27B success at
lower cost.  Bootstrap intervals based on paired tasks exclude zero for the
gain over SWE-Router in both model pairs.  Because \method{} invokes Strong on
most 4B tasks and 4B alone solves only two, this comparison says little about
settings in which occasional single-action assistance would suffice.

A larger study uses 1,000 training and 300 development tasks and evaluates
both model pairs on 274 tasks from the same underlying benchmark pool
(Appendix~\ref{app:alfworld-scaleup}). With 9B Cheap and 27B Strong, we can
ask whether the gain comes simply from preserving Cheap's progress. When
both policies continued from the current state, \method{} solved 159/274
tasks versus 137/274 for a fixed handoff after 16 Cheap actions, selected on
development data. Mean weighted call costs were 182.79 and 181.09,
respectively (paired gain: 8.03 percentage points; 95\% bootstrap interval
3.28--12.77). This comparison tests adaptive timing, not whether a
mixture-cure model is necessary. A discrete-time logistic hazard learned
from the same intervals as a matched \method{} refit attained similar
success at higher cost (Appendix~\ref{app:control-study}).

An ALFWorld trajectory illustrates what can be preserved. Asked to put a
clean ladle on a dining table, Cheap picked it up at action 2, then repeatedly
put it down and picked it up without washing it. Strong took over after
action 9 and finished in six actions; Cheap alone exhausted the budget.
To test the teacher's timing separately, we also handed off directly at
saved checkpoints. On 34 trajectories with all four comparison points,
Strong finished 21 tasks at the recommendation, compared with 23 three steps
earlier and 18 three steps later. These tasks were selected because Cheap
failed and Strong succeeded from the start; the results measure recovery
within that group, not how often an arbitrary task needs rescue
(Appendix~\ref{app:window-study}).

\subsection{DABench}

InfiAgent-DABench evaluates data-analysis agents in an executable tool
environment~\citep{hu2024dabench}.  We pair Qwen3.5-9B with Qwen3.6-27B, allow
at most 12 model calls, and use a fixed Train140/Dev50/Test67 split; full protocol details
appear in Appendix~\ref{app:protocol}.

On Test67, \method{} solved 49 tasks at a mean cost of 19.81; each of the
three routing baselines solved 45 at a higher mean cost. The shared call
budget includes Cheap proposals even when they are discarded at handoff.
The paired success interval versus SWE-Router includes zero, so the
four-task difference is suggestive rather than conclusive.

\FloatBarrier
\section{Conclusion and Limitations}
\label{sec:conclusion}

The contribution is a way to supervise handoff without running Strong from
every training checkpoint. A teacher can provide a useful interval without
knowing an exact switch time. A mixture-cure threshold model is one
interpretable way to learn from those labels; a discrete-time hazard can use
them as well. The real-task controls distinguish the value of choosing when
to hand off from the simpler benefit of continuing a partially completed task.

The score is not a causal estimate of rescue, and the teacher may mark a
window too early or too late. The timing controls are from ALFWorld, not
DABench. Our cumulative score cannot fall when an agent makes progress,
and a one-way handoff cannot exploit brief assistance followed by a return
to Cheap. Finally, a budget on weighted calls need not imply the same
ranking under token, latency, or monetary budgets. These choices should
depend on the environment and the cost the practitioner actually pays.

\subsection*{Reproducibility statement}

The appendices give the data splits, teacher-label encoding, fitting
objective, operating-point selection rules and evaluation costs. The
checkpoint comparisons report both the eligibility criteria and the number
of tasks tested at all offsets.

\subsection*{AI use statement}

OpenAI Codex assisted with programming, data analysis, and manuscript
editing. The primary experiments used DeepSeek V4 Pro for offline annotation;
Appendix~\ref{app:control-study} identifies the teacher used in the separate
timing-model comparison. The authors are responsible for the final content.

\bibliographystyle{preprint}
\bibliography{references}

\begin{thebibliography}{15}
\providecommand{\natexlab}[1]{#1}
\providecommand{\url}[1]{\texttt{#1}}
\expandafter\ifx\csname urlstyle\endcsname\relax
  \providecommand{\doi}[1]{doi: #1}\else
  \providecommand{\doi}{doi: \begingroup \urlstyle{rm}\Url}\fi

\bibitem[Chen et~al.(2024)Chen, Zaharia, and Zou]{chen2024frugalgpt}
Lingjiao Chen, Matei Zaharia, and James Zou.
\newblock {FrugalGPT}: How to use large language models while reducing cost and
  improving performance.
\newblock \emph{Transactions on Machine Learning Research}, 2024.
\newblock URL \url{https://openreview.net/forum?id=cSimKw5p6R}.

\bibitem[Farewell(1982)]{farewell1982mixture}
V.~T. Farewell.
\newblock The use of mixture models for the analysis of survival data with
  long-term survivors.
\newblock \emph{Biometrics}, 38\penalty0 (4):\penalty0 1041--1046, 1982.
\newblock \doi{10.2307/2529885}.

\bibitem[Heitjan \& Rubin(1991)Heitjan and Rubin]{heitjan1991coarse}
Daniel~F. Heitjan and Donald~B. Rubin.
\newblock Ignorability and coarse data.
\newblock \emph{The Annals of Statistics}, 19\penalty0 (4):\penalty0
  2244--2253, 1991.
\newblock \doi{10.1214/aos/1176348396}.

\bibitem[Hu et~al.(2024)Hu, Zhao, Wei, Chai, Ma, Wang, Wang, Su, Xu, Zhu,
  Cheng, Yuan, Li, Kuang, Yang, Yang, and Wu]{hu2024dabench}
Xueyu Hu, Ziyu Zhao, Shuang Wei, Ziwei Chai, Qianli Ma, Guoyin Wang, Xuwu Wang,
  Jing Su, Jingjing Xu, Ming Zhu, Yao Cheng, Jianbo Yuan, Jiwei Li, Kun Kuang,
  Yang Yang, Hongxia Yang, and Fei Wu.
\newblock {I}nfi{A}gent-{DAB}ench: Evaluating agents on data analysis tasks.
\newblock In \emph{Proceedings of the 41st International Conference on Machine
  Learning}, volume 235 of \emph{Proceedings of Machine Learning Research},
  pp.\  19544--19572. PMLR, 2024.

\bibitem[Kapoor et~al.(2026)Kapoor, Gupta, Chen, Beniwal, Huang, and
  Kumar]{kapoor2026trim}
Vansh Kapoor, Aman Gupta, Hao Chen, Anurag Beniwal, Jing Huang, and Aviral
  Kumar.
\newblock {TRIM}: Hybrid inference via targeted stepwise routing in multi-step
  reasoning tasks.
\newblock In \emph{International Conference on Learning Representations}, 2026.

\bibitem[Kuk \& Chen(1992)Kuk and Chen]{kuk1992mixture}
Anthony Y.~C. Kuk and Chen-Hsin Chen.
\newblock A mixture model combining logistic regression with proportional
  hazards regression.
\newblock \emph{Biometrika}, 79\penalty0 (3):\penalty0 531--541, 1992.
\newblock \doi{10.1093/biomet/79.3.531}.

\bibitem[Ong et~al.(2025)Ong, Almahairi, Wu, Chiang, Wu, Gonzalez, Kadous, and
  Stoica]{ong2025routellm}
Isaac Ong, Amjad Almahairi, Vincent Wu, Wei-Lin Chiang, Tianhao Wu, Joseph~E.
  Gonzalez, M.~Waleed Kadous, and Ion Stoica.
\newblock {RouteLLM}: Learning to route {LLM}s from preference data.
\newblock In \emph{International Conference on Learning Representations}, 2025.

\bibitem[Piatrashyn et~al.(2026)Piatrashyn, Kotelevskii, Grishchenkov, Glazkov,
  Nasonov, Makarov, Baldwin, Nakov, Vashurin, and Panov]{redact2026}
Dzianis Piatrashyn, Nikita Kotelevskii, Kirill Grishchenkov, Nikita Glazkov,
  Ivan Nasonov, Ilya Makarov, Timothy Baldwin, Preslav Nakov, Roman Vashurin,
  and Maxim Panov.
\newblock {ReDAct}: Uncertainty-aware deferral for {LLM} agents, 2026.
\newblock URL \url{https://arxiv.org/abs/2604.07036}.

\bibitem[Shridhar et~al.(2021)Shridhar, Yuan, C\^{o}t\'{e}, Bisk, Trischler,
  and Hausknecht]{shridhar2021alfworld}
Mohit Shridhar, Xingdi Yuan, Marc-Alexandre C\^{o}t\'{e}, Yonatan Bisk, Adam
  Trischler, and Matthew Hausknecht.
\newblock {ALFWorld}: Aligning text and embodied environments for interactive
  learning.
\newblock In \emph{International Conference on Learning Representations}, 2021.

\bibitem[Son et~al.(2026)Son, Yoon, Tang, Wang, Wolf, and
  Bogunovic]{son2026swerouter}
Seongho Son, Sangwoong Yoon, Jiahua Tang, Shuhan Wang, Lorenz Wolf, and Ilija
  Bogunovic.
\newblock {SWE-Router}: Routing in multi-turn agentic software engineering
  tasks, 2026.
\newblock URL \url{https://arxiv.org/abs/2607.00053}.

\bibitem[Spencer et~al.(2020)Spencer, Choudhury, Barnes, Schmittle, Chiang,
  Ramadge, and Srinivasa]{spencer2020interventions}
Jonathan Spencer, Sanjiban Choudhury, Matt Barnes, Matthew Schmittle, Mung
  Chiang, Peter Ramadge, and Siddhartha Srinivasa.
\newblock Learning from interventions: Human-robot interaction as both explicit
  and implicit feedback.
\newblock In \emph{Proceedings of Robotics: Science and Systems}, 2020.
\newblock \doi{10.15607/RSS.2020.XVI.055}.

\bibitem[Turnbull(1976)]{turnbull1976interval}
Bruce~W. Turnbull.
\newblock The empirical distribution function with arbitrarily grouped,
  censored and truncated data.
\newblock \emph{Journal of the Royal Statistical Society: Series B},
  38\penalty0 (3):\penalty0 290--295, 1976.
\newblock \doi{10.1111/j.2517-6161.1976.tb01597.x}.

\bibitem[Wei(1992)]{wei1992aft}
Lee-Jen Wei.
\newblock The accelerated failure time model: A useful alternative to the {Cox}
  regression model in survival analysis.
\newblock \emph{Statistics in Medicine}, 11\penalty0 (14--15):\penalty0
  1871--1879, 1992.
\newblock \doi{10.1002/sim.4780111409}.

\bibitem[Zhang et~al.(2026)Zhang, Yu, Yang, Wu, Huang, Li, and
  Yan]{zhang2026evoroute}
Guibin Zhang, Haiyang Yu, Kaiming Yang, Bingli Wu, Fei Huang, Yongbin Li, and
  Shuicheng Yan.
\newblock {EvoRoute}: Experience-driven self-routing {LLM} agent systems.
\newblock In \emph{Proceedings of the 64th Annual Meeting of the Association
  for Computational Linguistics (Volume 1: Long Papers)}, pp.\  38213--38225,
  San Diego, California, United States, July 2026. Association for
  Computational Linguistics.
\newblock \doi{10.18653/v1/2026.acl-long.1771}.

\bibitem[Zhang et~al.(2025)Zhang, Yin, Zhang, Liu, Han, Zhang, Li, Wang, Wang,
  Chen, and Wu]{zhang2025attribution}
Shaokun Zhang, Ming Yin, Jieyu Zhang, Jiale Liu, Zhiguang Han, Jingyang Zhang,
  Beibin Li, Chi Wang, Huazheng Wang, Yiran Chen, and Qingyun Wu.
\newblock Which agent causes task failures and when? {O}n automated failure
  attribution of {LLM} multi-agent systems.
\newblock In \emph{Proceedings of the 42nd International Conference on Machine
  Learning}, volume 267 of \emph{Proceedings of Machine Learning Research},
  pp.\  76583--76599. PMLR, 2025.
\newblock URL \url{https://proceedings.mlr.press/v267/zhang25cq.html}.

\end{thebibliography}

\appendix
\raggedbottom

\section{Finite-Sample Theory}
\label{app:finite-sample-proof}

\subsection{Clean target and local regularity}

Let \(O_i^0\) denote the clean observed TACIT record for task \(i\), including
its task features, risk-diagnostic sequence, paired-Strong outcome, and
interval or right-censoring information.  Let \(O_i\) be the record actually
used for fitting after possible teacher-interval corruption; all other fields
agree with \(O_i^0\).  Here, ``clean'' includes a valid coarsening whose
interval contains the latent crossing; replacing it with an invalid interval
is corruption.  Using a locally identifiable parameterization of
\((w,\gamma,\beta,s)\), let \(\ell(\theta;O)\) denote the negative
log-likelihood contribution in Equation~\eqref{eq:working-likelihood} and let
\(\mathcal P(\theta)\) denote the deterministic regularizer.  Define
\begin{align}
  M(\theta)
  &=\E[\ell(\theta;O^0)]+\mathcal P(\theta), \nonumber\\
  \widehat M_n^0(\theta)
  &=\frac1n\sum_{i=1}^n\ell(\theta;O_i^0)+\mathcal P(\theta), \qquad
  \widehat M_n(\theta)
  =\frac1n\sum_{i=1}^n\ell(\theta;O_i)+\mathcal P(\theta).
  \label{eq:clean-corrupted-objectives}
\end{align}
The simplex is represented with \(d-1\) independent coordinates rather than
all \(d\) softmax logits, which removes the common-shift nonidentifiability.
Any smooth local coordinate for the positive scale may be used.

\begin{assumption}[Local regularity and bounded teacher-interval influence]
\label{ass:local-regularity}
The clean records \(O_1^0,\ldots,O_n^0\) are i.i.d., and all recordwise losses
below are twice continuously differentiable on
the closed ball \(\mathbb B_r(\theta^\dagger)\), which is contained in the
interior of the local coordinate chart.  Differentiation and expectation may
be interchanged through order two on this ball.  There are constants
\(r,\mu,\sigma,G_T,H_T>0\) such that:
\begin{enumerate}[label=(\roman*),leftmargin=*,itemsep=2pt]
\item \(\theta^\dagger\) is an interior stationary point of \(M\), and \(M\)
is twice continuously differentiable with
\[
  \nabla^2M(\theta)\succeq\mu I_p
  \quad\text{for every }\theta\in\mathbb B_r(\theta^\dagger).
\]
\item The centered clean score
\[
  \xi_i=\nabla\ell(\theta^\dagger;O_i^0)
  -\E[\nabla\ell(\theta^\dagger;O_i^0)]
\]
is directionally \(\sigma\)-sub-Gaussian: for every unit vector \(v\) and
\(\lambda\in\mathbb R\),
\[
  \E\exp\{\lambda v^\top\xi_i\}
  \le \exp(\sigma^2\lambda^2/2).
\]
\item For the confidence level \(\delta\) under consideration, with
probability at least \(1-\delta/2\),
\[
  \sup_{\theta\in\mathbb B_r(\theta^\dagger)}
  \|\nabla^2\widehat M_n^0(\theta)-\nabla^2M(\theta)\|_{\rm op}
  \le\mu/4.
\]
\item Let \(\mathcal I_T\) denote the records with a replaced teacher
interval, where \(|\mathcal I_T|\le\rho_Tn\).  For every record,
\begin{align}
 \|\nabla\ell(\theta^\dagger;O_i)-\nabla\ell(\theta^\dagger;O_i^0)\|_2
 &\le G_T\ind\{i\in\mathcal I_T\}, \nonumber\\
 \sup_{\theta\in\mathbb B_r(\theta^\dagger)}
 \|\nabla^2\ell(\theta;O_i)-\nabla^2\ell(\theta;O_i^0)\|_{\rm op}
 &\le H_T\ind\{i\in\mathcal I_T\}.
 \label{eq:additive-corruption-influence}
\end{align}
Moreover, \(H_T\rho_T\le\mu/4\).
\end{enumerate}
\end{assumption}

For the log-normal working likelihood, sufficient local conditions for
item~(iv) are bounded covariates and diagnostics, a scale \(s\) bounded away
from zero, and interval and right-censor likelihood masses bounded away from
zero on \(\mathbb B_r(\theta^\dagger)\).  The local smoothness assumption also
requires the numerical likelihood floor to remain inactive on this ball.

\subsection{Proof of Theorem~\ref{thm:supervision-robustness}}

Let
\[
  Z_n=\frac1n\sum_{i=1}^n\xi_i.
\]
Choose a \(1/2\)-net \(\mathcal N\) of the Euclidean unit sphere in
\(\mathbb R^p\) with \(|\mathcal N|\le5^p\).  For any fixed
\(v\in\mathcal N\), the directional sub-Gaussian assumption gives
\[
  \Prob\!\left(\left|v^\top Z_n\right|>t\right)
  \le 2\exp\!\left(-\frac{nt^2}{2\sigma^2}\right).
\]
Taking
\[
  t=\sigma\sqrt{\frac{2\{p\log5+\log(4/\delta)\}}{n}}
\]
and applying a union bound over \(\mathcal N\) shows that, with probability
at least \(1-\delta/2\),
\[
  \max_{v\in\mathcal N}|v^\top Z_n|\le t.
\]
For any vector \(z\), the \(1/2\)-net property implies
\(\|z\|_2\le2\max_{v\in\mathcal N}|v^\top z|\).  Hence
\begin{equation}
  \|Z_n\|_2
  \le 2\sigma\sqrt{\frac{2\{p\log5+\log(4/\delta)\}}{n}}.
  \label{eq:score-concentration-proof}
\end{equation}

By Assumption~\ref{ass:local-regularity}(i),
\(\nabla M(\theta^\dagger)=0\).  The deterministic regularizer appears
identically in \(M\) and \(\widehat M_n^0\), so it cancels after centering and
\[
  \nabla\widehat M_n^0(\theta^\dagger)=Z_n.
\]
Let \(\mathcal I_T\) be the set of teacher-interval replacements.  The
recordwise gradient bound and the triangle inequality give
\begin{align}
 \|\nabla\widehat M_n(\theta^\dagger)\|_2
 &\le \|\nabla\widehat M_n^0(\theta^\dagger)\|_2
 +\frac{G_T|\mathcal I_T|}{n}\nonumber\\
 &\le
 2\sigma\sqrt{\frac{2\{p\log5+\log(4/\delta)\}}{n}}
 +G_T\rho_T
 =a_n(\delta).
 \label{eq:corrupted-gradient-proof}
\end{align}

On the clean empirical-Hessian event in
Assumption~\ref{ass:local-regularity}(iii),
\[
  \nabla^2\widehat M_n^0(\theta)\succeq(3\mu/4)I_p
  \quad\text{throughout }\mathbb B_r(\theta^\dagger).
\]
The recordwise Hessian bounds and item~(iv) imply
\[
  \sup_{\theta\in\mathbb B_r(\theta^\dagger)}
  \|\nabla^2\widehat M_n(\theta)
    -\nabla^2\widehat M_n^0(\theta)\|_{\rm op}
  \le H_T\rho_T\le\mu/4.
\]
Weyl's inequality therefore yields
\begin{equation}
  \nabla^2\widehat M_n(\theta)\succeq(\mu/2)I_p
  \quad\text{for all }\theta\in\mathbb B_r(\theta^\dagger).
  \label{eq:empirical-local-curvature}
\end{equation}

By continuity and compactness, \(\widehat M_n\) attains a minimum on the
closed ball.
For any \(u\) with \(\|u\|_2=r\), the fundamental theorem of calculus and
Equation~\eqref{eq:empirical-local-curvature} give
\begin{align}
 \left\langle\nabla\widehat M_n(\theta^\dagger+u),u\right\rangle
 &=
 \left\langle\nabla\widehat M_n(\theta^\dagger),u\right\rangle
 +\int_0^1
 u^\top\nabla^2\widehat M_n(\theta^\dagger+tu)u\,dt\nonumber\\
 &\ge -a_n(\delta)r+\frac{\mu r^2}{2}>0.
 \label{eq:outward-gradient}
\end{align}
Equation~\eqref{eq:outward-gradient} rules out boundary minimizers, so the
minimizer is interior and stationary.  Equation~\eqref{eq:empirical-local-curvature}
makes the objective \(\mu/2\)-strongly convex on the ball, hence the minimizer
is unique there.

Let \(\Delta=\widehat\theta-\theta^\dagger\).  Strong monotonicity of the
gradient and stationarity of \(\widehat\theta\) yield
\begin{align}
 \frac{\mu}{2}\|\Delta\|_2^2
 &\le
 \left\langle
 \nabla\widehat M_n(\widehat\theta)
 -\nabla\widehat M_n(\theta^\dagger),\Delta
 \right\rangle\nonumber\\
 &=
 -\left\langle\nabla\widehat M_n(\theta^\dagger),\Delta\right\rangle
 \le a_n(\delta)\|\Delta\|_2.
\end{align}
If \(\Delta\ne0\), division by \(\|\Delta\|_2\) proves
Equation~\eqref{eq:supervision-robustness}; the case \(\Delta=0\) is
immediate.

Finally, fix any evaluation task and decision step with features \(x\) and
observed diagnostics \(u(1{:}t)\).  The line segment between
\(\theta^\dagger\) and \(\widehat\theta\) remains in the ball, so the uniform
Lipschitz condition in the theorem gives
\[
 |q_{\widehat\theta}(x,u(1{:}t))
   -q_{\theta^\dagger}(x,u(1{:}t))|
 \le L_q\|\widehat\theta-\theta^\dagger\|_2
 \le\frac{2L_qa_n(\delta)}{\mu}.
\]
Because the task and step were arbitrary, this proves
Equation~\eqref{eq:score-recovery}.
The score-concentration and Hessian events each fail with probability at most
\(\delta/2\); their intersection therefore has probability at least
\(1-\delta\), completing the proof.

\section{Protocol Details}
\label{app:protocol}

\subsection{Controlled Statistical Protocols}

\paragraph{Model-based AB--AFT study.}
The recovery experiment uses 500 independent fits at each
\(n\in\{500,2{,}000,8{,}000\}\), three optimizer restarts, at most 500
iterations, and a fresh 10,000-task evaluation sample.  The matched-noise
experiment uses 100 paired replicates, each with 4,000 training and 4,000 test
tasks, two restarts, and at most 350 iterations.  Both experiments fix the
scalar risk weight at one and use no regularization.  For \(AB=11\), they use
the right-censoring contribution with observed \(B_i=1\).  Pilot calibration
and evaluation use disjoint random streams; every fit trained with corrupted
supervision is evaluated on clean held-out data.

\paragraph{Mechanism-based data generation and supervision.}
Each task has ten opportunities and succeeds after at least eight correct
actions.  Its execution state is on-track, recoverable, or off-track.  Cheap
and Strong follow fixed state-transition and action-success probabilities,
with Strong more likely to recover from degraded states.  A post-step Cheap
risk signal provides partial evidence about the realized path and can affect
routing only from the following step onward.

On each Pure-Cheap trajectory, the simulator searches post-action checkpoints
1--5 and records the earliest candidate handoff checkpoint at which the task
remains unfinished, cumulative Cheap risk is at or above the task-specific
threshold, the resulting state is recoverable or off-track, and Strong's
probability of a correct next action exceeds Cheap's by at least \(0.08\).
Under operational AB supervision, this candidate is retained only for
\(AB=01\).  A retained candidate becomes a one-step interval at that
post-action checkpoint.  An \(AB=01\) record without a candidate contributes
only its observed \(B_i=1\) status; \(AB=11\) is right-censored at the horizon,
and \(AB\in\{00,10\}\) enters the cure component.  The latent threshold and
candidate checkpoint do not enter actor dynamics, risk emissions, or terminal
success.

Each reported replicate contains 8,000 training, 4,000 development, and 20,000
test tasks.  The main comparison and feature ablation use
100 replicates, while each nonbaseline sensitivity setting uses 50.  All
methods use costs \(c_C=1,c_S=3\) and the common development cap 16.
Pilot calibration and reported evaluation use disjoint random streams.  The
sensitivity experiments vary only learner-visible task features or
risk-emission noise, leaving actor dynamics and coupled endpoint outcomes
unchanged.

\paragraph{Compared policies.}
We compare four learned policies on the same coupled tasks.  Task Router
estimates the probability of Pure-Cheap success from task features and chooses
Cheap or Strong before execution.  Step Deferral uses the latest observed Cheap risk to assign only
the following step to Strong, after which control returns to Cheap.  The
SWE-Router analogue predicts Pure-Cheap success from task features and the
first \(K\in\{1,2,3,4\}\) Cheap risk observations, then decides either to
continue with Cheap or to restart with Strong.  \method{} combines task features
with cumulative Cheap risk and, at its first score crossing, hands control
permanently to Strong from the current state.  At deployment, none of these
policies observes latent states, teacher labels, Strong outcomes, or terminal
rewards.  We use 101 quantiles of the training scores as candidate thresholds;
each policy is
selected on development under the common cost cap and frozen before test,
with \(K\) and the threshold selected jointly for the fixed-prefix policy.
The SWE-Router baseline is a structural analogue because the simulator
provides numerical features rather than language-model trajectories.  Its
fixed-\(K\) diagnostics use the same train/development/test splits.

Diagnostic checks show that cumulative risk predicts terminal Pure-Cheap
failure better than task features alone or the latest risk value (AUC
\(.774\) versus \(.605\) and \(.691\)).  Across the 100 held-out replicates,
the mean fraction of \(AB=01\) tasks with a candidate handoff checkpoint is
\(80.1\%\).  Among those tasks, permanently handing off immediately after the
marked Cheap step produces terminal success in \(85.6\%\) of cases.

\paragraph{Sensitivity results.}
Across all sensitivity settings, \method{} has higher held-out success than
the best routing baseline selected on development data at comparable cost, with larger
gains as online-risk information becomes more informative.

\subsection{Interactive-Agent Protocols}

\paragraph{Interactive-agent cost accounting.}
For reproducibility, the reported proxy uses common 4B-equivalent units.  If
Cheap model \(m\in\{4\mathrm{B},9\mathrm{B}\}\) makes \(n_{m,i}\) calls, the
27B Strong model makes \(n_{27,i}\) calls, and the 7B value head is invoked
\(h_i\in\{0,1\}\) times, then
\begin{equation}
  \mathrm{Cost}_i=c_m n_{m,i}+6.75n_{27,i}+1.75h_i,\qquad
  c_{4\mathrm{B}}=1,\;c_{9\mathrm{B}}=2.25.
  \label{eq:cost}
\end{equation}
The proxy selects the development operating point and reports realized test
cost; it does not enter score fitting.  SWE-Router includes its Cheap prefix,
one head call when invoked, and the full Strong restart.  Offline teacher
labeling is excluded from deployment cost.

\paragraph{ALFWorld.}
The fit/select/test split is \textsc{seen100}/\textsc{seen40}/
\textsc{unseen134}.  For the 4B--27B pair, the training counts in
\((AB=00,01,10,11)\) order are \((34,58,1,7)\); for 9B--27B they are
\((35,43,0,22)\).  Every \(AB=01\) training episode receives a teacher
interval.  The task router evaluates thresholds
\(\{0,0.05,0.10,\ldots,0.95,1.01\}\); selected-action-NLL deferral targets
an average of \(k\in\{0,5,10,15,20,25,30,35,40\}\) Strong-deferred steps
per training task; and \method{} uses
\(\alpha\in\{0.1,0.2,\ldots,0.9\}\).  Selection maximizes
\textsc{seen40} success subject to at most 90\% of pure-27B proxy cost, breaking
ties by lower cost.  Development selects \(\alpha=0.2\) for 4B and
\(\alpha=0.3\) for 9B; both values are frozen before \textsc{unseen134}
evaluation.

\paragraph{DABench.}
We use a fixed custom split of a 257-task DAEval subset into Train140, Dev50,
and Test67, with Test67 held out from all fitting and development selection.
We do not present this subset or split as an official benchmark partition.
The task-router target rates are \(\{0,0.125,\ldots,0.875,1\}\), where each
rate targets that fraction of training tasks for Strong execution.
Selected-action-NLL deferral targets an average of
\(k\in\{0,0.75,1.5,2.25,3,3.75,4.5,5.25,6\}\) deferred steps per training
task.  For \method{}, a deterministic grid combines target train-switch rates
with train-score quantiles in \(\{0,0.1,\ldots,1\}\).  Using training data
only, we deduplicate candidates by their induced policies and thin them to
nine points, including both endpoints.  Dev50 then selects the highest-success
candidate using at most 80\% of pure-27B proxy cost; remaining ties are broken
by lower cost, intervention rate, and target rate.  The selected settings are
a 62.5\% Task Router target, \(k=1.5\), and a 40\% \method{} train-switch
target corresponding to \(\alpha\approx0.163\).  The nine prespecified
\method{} candidates are shown on Test67 only as a diagnostic curve; the
reported result uses the development-selected 40\% candidate.

\paragraph{Compared policy implementations.}
Task Router fits a logistic model for Pure-Cheap success from task-level
features and chooses Cheap or Strong before execution.  Step Deferral
thresholds the Cheap model's selected-action negative log-likelihood and
delegates only the current action to Strong.  The SWE-Router-style baseline
fine-tunes a Qwen2.5-Coder-7B-Instruct value head with LoRA to predict
Pure-Cheap success from the task and the visible Cheap prefix at
\(K\in\{1,2,3,4\}\); it then decides either to continue with Cheap or to
restart with Strong.  \method{} fits the censored threshold model described in
Appendix~\ref{app:objective} and permanently hands off from the current state.
All model fitting uses training data, and every policy parameter is selected
on development under the benchmark cost cap and frozen before test.  The
fixed-prefix policy never receives Strong outcomes, teacher labels, or
\method{} scores as inputs, and its Cheap prefix, value-head call, and Strong
restart are all charged by Equation~\eqref{eq:cost}.  Because the available
ALFWorld cache omits hidden Cheap reasoning and the original paraphrase
augmentation is unavailable, we call it a SWE-Router-style reproduction.

For the selected operating points, paired task-level bootstrap intervals use
10,000 with-replacement resamples and percentile endpoints.  The 95\% intervals
for the success-rate difference between \method{} and SWE-Router are
\([17.2,35.1]\) percentage points for ALFWorld 4B--27B,
\([2.2,17.9]\) for ALFWorld 9B--27B, and \([-3.0,14.9]\) for DABench 9B--27B.

\paragraph{Offline teacher annotation.}
For each \(AB=01\) training episode, an offline teacher observes the task,
the failed Cheap trajectory, and the successful Strong trajectory.  It returns
the earliest range of post-action checkpoints at which failure risk is clear and
recovery remains feasible, together with a recommended checkpoint in that
range.  A mark at step \(t\) denotes handoff after observing Cheap's action and
result at step \(t\) and before step \(t+1\).  The primary experiments treat the
recommended checkpoint as the one-step interval \([t,t]\), whereas the larger
ALFWorld study retains the complete range.  Teacher explanations are not used
for fitting.

\paragraph{Decision timing.}
ALFWorld executes the Cheap action, observes the resulting state, updates risk,
and, if the threshold is crossed, gives the next action to the Strong agent.
DABench instead scores an unexecuted Cheap proposal.  If the threshold is
crossed, it discards that proposal and asks the Strong agent to act from the
prior tool history; otherwise it executes the proposal.  For DABench, a teacher
mark after step \(t\) is aligned with the proposal at step \(t+1\) to match this
pre-execution decision point.

\FloatBarrier

\subsection{Larger-Sample ALFWorld Scale-Up Evaluation}
\label{app:alfworld-scaleup}

We use this study as a scale check rather than as a separate benchmark.  The
fit and development sets contain 1,000 and 300 episodes from the official training split
from disjoint parent-task groups; groups overlapping the official validation
sets are excluded.  Test274 is the union of all 140 valid-seen and 134
valid-unseen tasks.  In the 4B--27B training set, the counts for
\(AB=00,01,10,11\) are \(371/576/2/51\); in the 9B--27B set, they are
\(351/444/22/183\).
DeepSeek V4 Pro labels every AB01 training trajectory.  Unlike the primary
ALFWorld fit, the scale-up fit uses the complete teacher interval
\([\tau_{\rm low},\tau_{\rm high}]\); most of these intervals span multiple
steps.

All candidates are selected on Dev300 under the same cost cap, \(180.77\)
4B-equivalent units, and then frozen.  Development selects
\(\alpha=0.3\) for 4B--27B and \(\alpha=0.2\) for 9B--27B; the SWE-Router-style
baseline selects \(K=4\) in both pairs.  Test costs are computed from actual
model-call counts under Equation~\eqref{eq:cost}.

\begin{figure}[H]
\centering
\includegraphics[width=\linewidth]{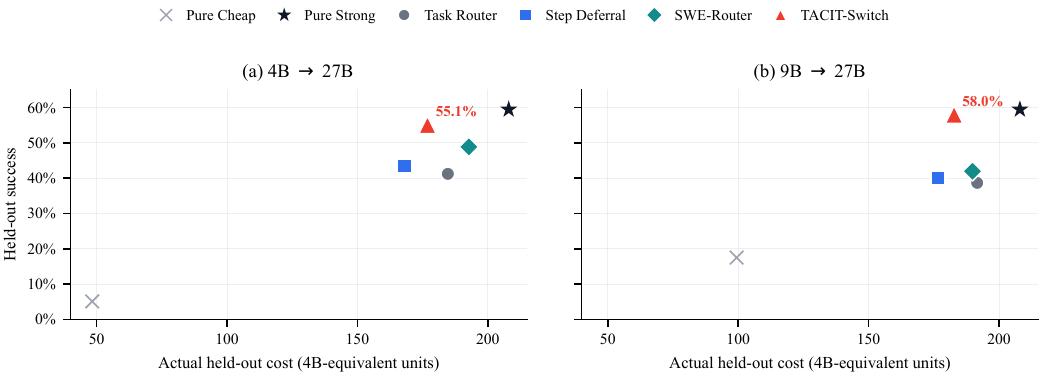}
\caption{\textbf{Larger-sample ALFWorld success--cost operating points.}
The displayed policies were frozen on Dev300 and evaluated on the same
Test274 tasks. Costs follow Equation~\eqref{eq:cost} using model calls.}
\label{fig:alfworld-scaleup-operating}
\end{figure}

Figure~\ref{fig:alfworld-scaleup-operating} shows that \method{} has the
highest success point estimate among learned policies in both model pairs.
It is also less costly than Task Router and SWE-Router.  Relative to Step
Deferral, it costs 8.8 additional units and gains 11.7 percentage points for
4B--27B.  For 9B--27B, the corresponding differences are 6.1 units and 17.9
percentage points.

\begin{table}[H]
\caption{\textbf{Larger-sample ALFWorld operating points on Test274.}
Policies are selected on Dev300.  Intervention is the fraction of tasks that
invoke Strong; Strong share is the fraction of all model calls made by Strong.  \(\dagger\)
marks the proposed method.}
\label{tab:alfworld-scaleup-operating}
\centering
\scriptsize
\setlength{\tabcolsep}{3.2pt}
\begin{tabular}{llrrrr}
\toprule
Pair & Policy & Success & Avg. cost & Strong share & Intervention \\
\midrule
\multirow{6}{*}{4B--27B}
 & Pure 4B & 14/274 (5.1\%) & 48.32 & 0.0\% & 0.0\% \\
 & Task Router & 113/274 (41.2\%) & 184.69 & 70.0\% & 75.5\% \\
 & Step Deferral & 119/274 (43.4\%) & 168.05 & 35.0\% & 100.0\% \\
 & SWE-Router (\(K=4\)) & 134/274 (48.9\%) & 192.75 & 70.5\% & 81.8\% \\
 & \method{}\(^{\dagger}\) & 151/274 (55.1\%) & 176.87 & 79.1\% & 95.6\% \\
 & Pure 27B & 163/274 (59.5\%) & 208.02 & 100.0\% & 100.0\% \\
\midrule
\multirow{6}{*}{9B--27B}
 & Pure 9B & 48/274 (17.5\%) & 99.40 & 0.0\% & 0.0\% \\
 & Task Router & 106/274 (38.7\%) & 191.64 & 63.3\% & 63.5\% \\
 & Step Deferral & 110/274 (40.1\%) & 176.69 & 26.9\% & 98.9\% \\
 & SWE-Router (\(K=4\)) & 115/274 (42.0\%) & 189.85 & 55.9\% & 61.3\% \\
 & \method{}\(^{\dagger}\) & 159/274 (58.0\%) & 182.79 & 71.0\% & 92.7\% \\
 & Pure 27B & 163/274 (59.5\%) & 208.02 & 100.0\% & 100.0\% \\
\bottomrule
\end{tabular}
\end{table}

\FloatBarrier

\subsection{Progress-preserving routing controls}
\label{app:control-study}

To separate the choice of \emph{when} to hand off from the choice to preserve
progress, we compared adaptive handoff with a policy that always hands off
after a fixed number of Cheap actions if the task is still running. Both let
Strong finish from the state Cheap reached. Of the candidate delays (4, 8,
and 16 actions), we selected 16 on Dev300 by maximizing success under the
same cost budget used in the scale-up study. On the original 9B--27B
executions, adaptive handoff solved 22 more tasks at similar mean cost
(Table~\ref{tab:handoff-controls}); the paired success difference was 8.03
percentage points (95\% task-bootstrap interval: 3.28--12.77).

\begin{table}[htbp]
\centering
\caption{ALFWorld 9B--27B controls on the original 274 evaluation tasks.
Cost is the mean parameter-weighted call count. The three SWE-style policies
use the same four-action prefix and scoring head.}
\label{tab:handoff-controls}
\begin{tabular}{lrr}
\toprule
Policy & Tasks solved & Mean cost \\
\midrule
Task Router, matched 12 initial features & 127 & 191.88 \\
Fixed handoff after 16 actions & 137 & 181.09 \\
SWE-style restart, original threshold & 115 & 189.85 \\
SWE-style continuation, same threshold & 106 & 181.37 \\
SWE-style continuation, threshold reselected on Dev & 147 & 194.83 \\
\method{} & 159 & 182.79 \\
\bottomrule
\end{tabular}
\end{table}

Preserving progress alone did not improve the SWE-style router at its original
threshold; reselecting the threshold on Dev raised its success count to 147.
Giving the task-level router the same 12 initial features as \method{} (instead
of five) raised its 9B success count from 106 to 127. With 4B Cheap, the same
change reduced its count from 113 to 106 (cost 187.30), versus 151 at 176.87
for \method{}. Matching initial features does not give a pre-execution router
access to later trajectory diagnostics. A stepwise controller given all 12
task features and five trajectory diagnostics but delegating only one action
at a time solved 87/274 at cost 143.37 with 9B Cheap. Its action scoring
also differed, so this is not a test of permanence alone.
These comparisons reuse the scale-up test tasks. The original records lacked
action uncertainty on 174 tasks, for which the missing signal was set to
zero. In a separate 9B rerun with complete diagnostics, adaptive handoff solved more tasks than
fixed handoff (153 vs.\ 141), but at higher cost (192.77 vs.\ 178.96).

\paragraph{Changing the timing model, not the supervision.}
In a 200-task training subset, we obtained 89 new windows from the
\texttt{deepseek-flash} teacher, allowing wide windows when an exact step
was uncertain. With the same timing windows for both fits, we
kept the fitted Strong-success component fixed and replaced the threshold
distribution with a discrete-time logistic hazard. It uses task features,
elapsed time and trajectory diagnostics to estimate a stepwise hazard; the
likelihood assigns probability to the whole teacher interval. With thresholds
chosen on Dev300 under the same budget, the hazard solved 151/274 at mean
cost 199.56, versus 148/274 at 188.38 for the \method{} refit. Fitting the
same hazard to the teacher's recommended step alone solved 150/274 at 201.97.
Thus the timing model can be replaced without relabeling; this comparison
does not establish superiority of the mixture-cure form or of interval over
point labels in task success.

\subsection{Handoff timing on real trajectories}
\label{app:window-study}

\paragraph{What happens before handoff?}
Of 274 ALFWorld 9B Cheap-only runs, 237 repeated an action three times in a
row and 245 repeated a short cycle. Among runs exhibiting each pattern, its
first occurrence was at action 8.4 and 11.2 on average, respectively; the
groups overlap, and repetition alone need not indicate failure.
\method{} handed off on 254 of the 274 tasks, after a median of eight Cheap
actions (interquartile range 6--13). Relative to letting Cheap finish, these
handoffs recovered 113 failures but also turned two successes into failures.

\paragraph{Can Strong recover at a teacher-marked checkpoint?}
We identified 75 tasks where Cheap failed and Strong succeeded from the start,
with saved checkpoints at and three actions on either side of the teacher's
recommendation. Only 34 also had a checkpoint six actions earlier. For each of
these 34 tasks we ran Strong from all four states on the same Cheap trajectory
(Table~\ref{tab:teacher-checkpoints}). In the larger, three-checkpoint group,
Strong solved 56/75 three actions early, 52/75 at the recommendation and
47/75 three actions late. In a separate check on 50 eligible unseen tasks,
Strong solved 38 just before the window, 37 at the recommendation and 37
just after. The windows often contained recoverable states, but neighboring
checkpoints could work as well. Conversely, on 65 tasks where both models failed from the start, Strong
solved 12 after a fixed handoff at action 16. Initial Strong success is thus
neither necessary nor sufficient for rescue at a particular checkpoint.

\begin{table}[htbp]
\centering
\caption{\textbf{Direct handoffs on 34 matched ALFWorld tasks (9B--27B).}
Offsets count Cheap actions relative to the teacher's recommended checkpoint.
Mean cost includes the Cheap prefix and Strong continuation (Eq.~\eqref{eq:cost}).}
\label{tab:teacher-checkpoints}
\begin{tabular}{lrrrr}
\toprule
Offset in actions & $-6$ & $-3$ & $0$ & $+3$ \\
\midrule
Tasks solved (of 34) & 20 & 23 & 21 & 18 \\
Mean cost & 202.83 & 189.13 & 191.32 & 191.32 \\
\bottomrule
\end{tabular}
\end{table}

\paragraph{Calibrating the learned policy.}
On 117 annotated test tasks the fitted policy switched 2.31 actions later than
the teacher's recommendation on average. This discrepancy motivates checking
calibration; it cannot explain the table, where handoff times were set directly.
We separately shifted the \emph{training} intervals three actions later and
refit the model, retaining paired outcomes even when a shifted interval became
unusable. At the original threshold ($0.20$), Dev300 successes fell from
189 to 168 and mean cost rose from 180.05 to 184.06. Thus the fit was sensitive
to the timing information, not just to the paired outcomes. Jointly selecting
the interval shift ($-5$ to $+3$ actions) and threshold on all 300 development
tasks under the original cost cap selected the unshifted intervals and threshold
$0.20$. The average lag alone is not a reason to shift every label.

\FloatBarrier

\subsection{Feature Definitions and Optimization}
\label{app:features}

Task features are standardized using training-split means and scales; risk
diagnostics are used directly as nonnegative values.  No development or test
outcome enters feature scaling or fitting.

\paragraph{Baseline features and fitting.}
The ALFWorld task router uses five binary descriptors: whether the final
receptacle is openable, whether it is a surface, whether the task needs an
external tool, whether the requested state is cold, and whether the target is
food.  It fits
an L2-penalized logistic model by Newton updates (\(\lambda=1\), at most 100
iterations, tolerance \(10^{-9}\)).  The DABench task router instead uses
operation breadth and a transformation-workflow indicator, fitted by
L2-logistic regression with \(C=0.1\), LBFGS, and at most 2000 iterations.
The step-deferral baseline uses only selected-action NLL and the
benchmark-specific \(k\) grids stated above.

\paragraph{ALFWorld \method{} features.}
The 12-dimensional task vector encodes requested state, task form, target
category, receptacle type, and the visible-entity count.  Five step-level
diagnostics capture repeated actions, short action cycles, selected-action
uncertainty \(1-\exp(-\mathrm{NLL})\), invalid generations or fallbacks, and
unchanged observations with zero reward.  Candidate margin is not used, and missing
selected-action uncertainty is mapped to zero.  Joint fitting uses Adam with
five restarts, \(\lambda=0.01\), and early stopping.

\paragraph{DABench \method{} features.}
The two task features are normalized answer arity and an indicator that the
required output specifies decimal precision.  Eight diagnostics computed from
each proposal capture format or protocol violations, execution failures,
repetition, stalled answers, pressure from the remaining tool budget, lack of
recent progress, and semantic mismatch with the task.  Joint fitting uses Adam with six restarts, \(\lambda=0.03\), and
early stopping.

\FloatBarrier
\subsection{Deployment cost and sensitivity to training inputs}
\label{app:runtime-study}

\paragraph{Handoff context and cost.}
In ALFWorld, Strong receives the current observation, task, admissible actions,
and up to 4,000 characters of action--observation history, not Cheap's private
reasoning. In DABench it inherits the tool state and a bounded execution log,
but not the discarded proposal. Each proposal counts against the shared
12-call budget, even if not executed. On the 67 DABench test tasks, the
selected policy made 236 Cheap and 118 Strong calls: Strong's share was
33.3\%, and the mean cost in Equation~\eqref{eq:cost} was 19.81.
Neither monitor calls another model; for example, DABench flags semantic
mismatches by checking proposed code against schema, type, and operator rules.

Counting input and output tokens, including repeated history prefills,
\method{} averaged 9,893 tokens per DABench task. On the original 274-task
ALFWorld 9B--27B evaluation, the reconstructed means were 40,701 for \method{}
and 48,455 for SWE-Router with restart (including its value head).
Some output lengths were estimated by retokenization. Weighting tokens by
model size reversed the ordering: \method{} used 2.49\% more. Neither measure
is a wall-clock or monetary cost.

\paragraph{Diagnostic design and sensitivity.}
Feature design drew on an early ALFWorld seen-task pilot and DABench tool
errors explored before the final split; final weights were fitted on Train and
thresholds selected on Dev. In the original ALFWorld 9B scale-up fit, weights
for repetition, short cycles, action uncertainty, invalid actions, and
stagnation were 0.0002, 0.0007, 0.4915, 0.3125, and 0.1952, respectively.
In a separate refit with teacher windows shifted one step earlier throughout,
the full model solved 157/274 tasks. Omitting each diagnostic in that order
gave 157, 157, 159, 157, and 155 successes. No single omission cost more than
two successes under this protocol.

\paragraph{Sensitivity to the Strong outcome.}
On 100 ALFWorld training tasks, greedy Strong solved 55; three sampling runs
(temperature 0.7, top-$p$ 0.9) solved 60, 58, and 64, disagreeing on 26 tasks.
Replacing those 100 binary outcomes by their three-run means and refitting
only incidence left the other 900 labels, timing model, and gate fixed.
Handoff timing changed on 47/300 development tasks. The same 13 never
switched; among the remaining 287, the mean absolute change was 0.21 actions.
This tests handoff sensitivity, not end-to-end success across seeds.

\FloatBarrier
\section{Robustness to Noisy Teacher Supervision}
\label{app:noise-robustness}

Theorem~\ref{thm:supervision-robustness} covers incorrect teacher intervals
only.  As an additional empirical stress test, we also corrupt the observed
paired Strong outcome while keeping evaluation clean.  This study fixes
\(n=4{,}000\) and uses 100 paired replicates per condition.  Among \(AB=01\)
records with a one-step interval, teacher-interval noise moves a fraction
\(\rho_T^{\rm timed}\) to an adjacent step.  Strong-outcome noise flips \(B_i\) for a fraction \(\rho_C\)
of tasks.  The matched joint conditions set
\(\rho_T^{\rm timed}=\rho_C\), and no test measurement is corrupted.  If
\(n_T\) of the \(n\) records are timed, the record-level corruption fraction
in Theorem~\ref{thm:supervision-robustness} is
\(\rho_T=(n_T/n)\rho_T^{\rm timed}\).

\begin{figure}[H]
\centering
\includegraphics[width=0.72\textwidth]{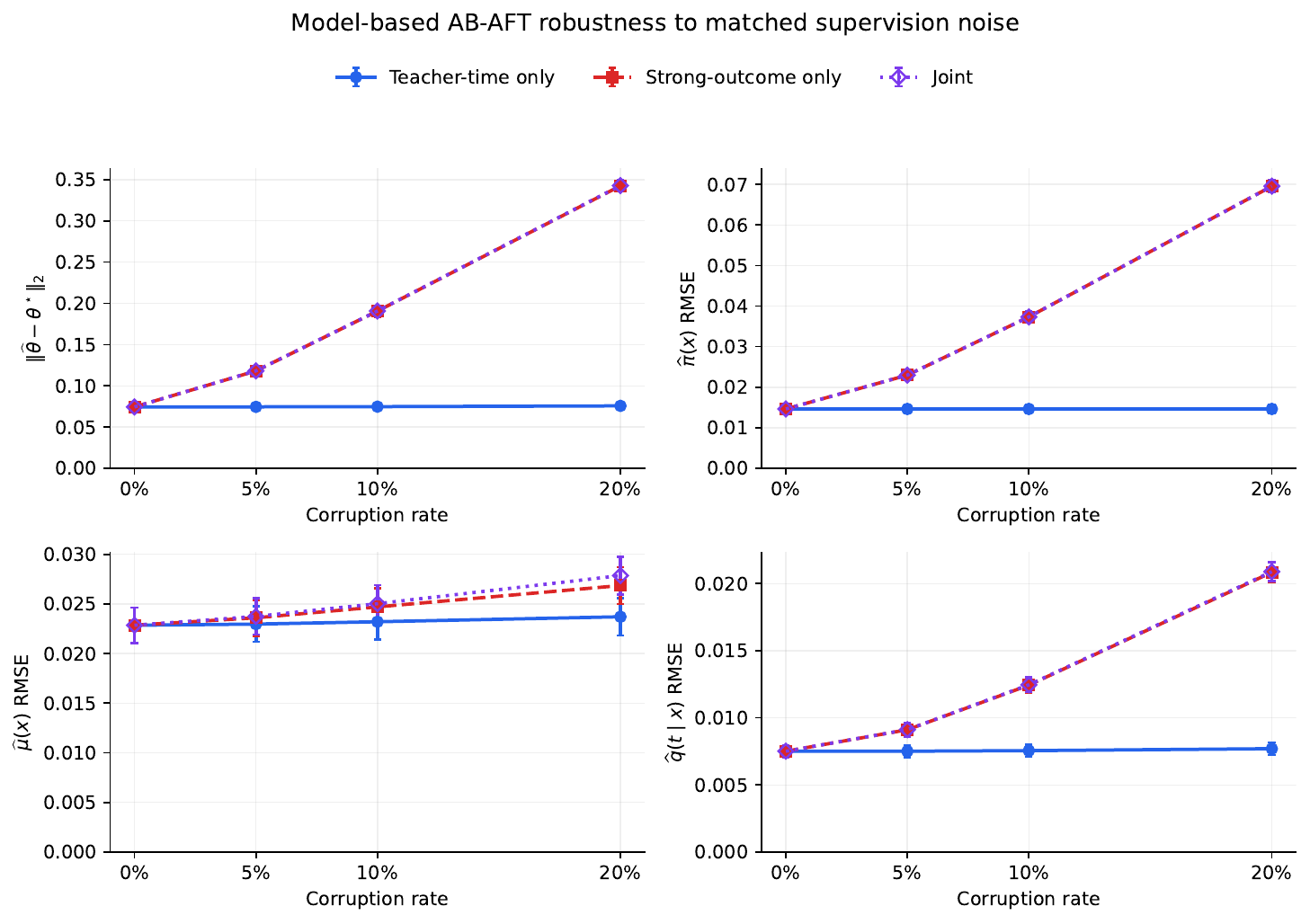}
\caption{Matched teacher-interval and Strong-outcome noise
under the correctly specified model-based DGP.  Points show mean estimation
or held-out prediction error over 100 paired replicates; bars show Student-\(t\)
95\% Monte Carlo confidence intervals for the replicate means.  The horizontal
coordinate is the direct corruption rate; for the joint condition, both noise
rates equal the displayed value.}
\label{fig:matched-noise}
\end{figure}

Clean \(q(t\mid x)\) RMSE is \(0.0075\pm0.0005\).  At 20\% corruption, it is
\(0.0077\pm0.0005\) under teacher-interval noise,
\(0.0208\pm0.0007\) under Strong-outcome noise, and
\(0.0209\pm0.0007\) under matched joint noise.
Figure~\ref{fig:matched-noise} plots these errors against the direct corruption
rate; for the joint condition, both \(\rho_T^{\rm timed}\) and \(\rho_C\)
equal the displayed rate.  Under this design, corrupting the paired Strong
outcome has a larger effect than shifting the teacher interval by one step.

\FloatBarrier

\clearpage
\section{Working Objective and Inference}
\label{app:objective}
\label{app:likelihood}

The implementation parameterizes \(w=\operatorname{softmax}(\widetilde w)\)
and uses a softplus transform to enforce \(s>0\).  With the per-episode
contributions in Equation~\eqref{eq:working-likelihood}, it applies a hard
numerical floor \(\varepsilon=10^{-12}\) to interval masses and final
per-record likelihoods.  Let \(\overline{\mathcal L}_i\) denote the resulting
implemented contribution.  The optimized objective is
\begin{equation}
 -\frac{1}{n}\sum_{i=1}^n\log\overline{\mathcal{L}}_i
 +\lambda\!\left(\lVert\gamma_{-0}\rVert_2^2+
                   \lVert\beta_{-0}\rVert_2^2\right).
 \label{eq:penalized-objective}
\end{equation}
The intercepts, scale, and risk-weight logits are unpenalized.  We optimize all
parameters jointly and retain the restart with the lowest objective value.  For
the parameterization used in Theorem~\ref{thm:supervision-robustness}, we fix
one logit to remove the softmax common-shift nonidentifiability and assume that
the numerical floors remain inactive in the local neighborhood.

At inference, both benchmarks update cumulative risk and evaluate
Equation~\eqref{eq:decision-rule} once per cheap-agent checkpoint;
Algorithm~\ref{alg:inference} specifies the benchmark-dependent checkpoint
semantics.

\begin{algorithm}[H]
\small
\caption{Benchmark-aware permanent handoff}
\label{alg:inference}
\begin{algorithmic}[1]
\Require fitted parameters, threshold \(\alpha\), cheap policy \(\cheap\),
strong policy \(\strong\)
\State \(R\gets0\); controller \(\gets\cheap\)
\While{episode is unfinished and budget remains}
  \If{controller is \(\strong\)}
    \State execute one strong-agent action and continue
  \EndIf
  \State obtain cheap checkpoint evidence \(e_t\)
  \Comment{post-action in ALFWorld; proposal in DABench}
  \State \(R\gets R+\hat w^\top u(e_t)\);
  \(q\gets\hat\pi F_{\hat\beta,\hat s}(R\mid x)\)
  \If{\(q\ge\alpha\)}
    \State controller \(\gets\strong\) permanently
    \If{\(e_t\) is an unexecuted proposal}
      \State discard the proposal; take a Strong action if the shared budget permits
    \EndIf
  \Else
    \State execute \(e_t\) if it is an unexecuted proposal
  \EndIf
\EndWhile
\end{algorithmic}
\end{algorithm}

\end{document}